\documentclass{article}

\usepackage{graphicx}
\usepackage{subfigure}
\usepackage{booktabs} 
\usepackage{tikz}
\usepackage{hhline}
\usetikzlibrary{shapes.geometric,arrows,positioning}

\graphicspath{{pics/}}

\usepackage{hyperref}

\usepackage[accepted]{icml2020}

\usepackage{mathrsfs,amsmath,amsfonts,amssymb,bm,bbm,dsfont,pifont,amscd,
stmaryrd,euscript,amsthm,appendix,color,epsfig,xr, authblk}

\def\W{\mathcal{W}}

\def\X{\mathcal{X}}

\def\Y{\mathcal{Y}}

\def\R{\mathbb{R}}
\def\C{\mathcal{C}}

\def\A{\mathcal{A}}

\def\P{\mathbb{P}}

\newcommand{\Acc}{\mathrm{Acc}\,}
\newcommand{\CM}{\C_{\mathrm{M}}}
\newcommand{\CF}{\C_{\mathrm{F}}}
\newcommand{\CS}{\C_{\mathrm{S}}}
\newcommand{\CC}{\C_{\mathrm{C}}}
\newcommand{\wstats}{\widetilde{W}_{\!L}}

\def\P{\mathbb{P}}

\begin{document}
\hyphenation{pa-ra-me-ter pa-ra-me-ters hy-per-pa-ra-me-ter hy-per-pa-ra-me-ters de-no-ted in-spi-ra-tion un-der-stan-ding con-tri-bu-tions Alex-ey Do-so-vits-kiy Alex-an-der Ko-les-ni-kov Lu-cas Beyer}
\twocolumn[
\icmltitle{Predicting Neural Network Accuracy from Weights}




\begin{icmlauthorlist}
\icmlauthor{Thomas Unterthiner}{goo}
\icmlauthor{Daniel Keysers}{goo}
\icmlauthor{Sylvain Gelly}{goo}
\icmlauthor{Olivier Bousquet}{goo}
\icmlauthor{Ilya Tolstikhin}{goo}
\end{icmlauthorlist}

\icmlaffiliation{goo}{Google Research (Brain Team)}

\icmlcorrespondingauthor{}{unterthiner@google.com}
\icmlcorrespondingauthor{}{tolstikhin@google.com}

\icmlkeywords{Machine Learning, ICML}

\vskip 0.3in
]


\printAffiliationsAndNotice{}  

\begin{abstract}
We show experimentally that the accuracy of a~trained neural network can be predicted surprisingly well by looking only at its weights, without evaluating it on input data. We motivate this task and introduce a formal setting for it. Even when using simple statistics of the weights, the predictors are able to rank neural networks by their performance with very high accuracy ($R^2$ score more than 0.98). Furthermore, the predictors are able to rank networks trained on different, unobserved datasets and with different architectures. 
We release a collection of 120k convolutional neural networks trained on four different datasets to encourage further research in this area, with the goal of understanding network training and performance better.
\end{abstract}

\section{Introduction}
Deep neural networks (DNNs) are considered state of the art methods for many machine learning problems today.
Yet, a deeper understanding of the mechanisms underlying these successes is still lacking.
The \emph{deep learning phenomena}, i.e.\ various surprising and insightful empirical findings surrounding the efforts to understand DNN training and generalization have recently gained a lot of attention from researchers and practitioners \cite{Zhang17,Frankle2019,Zhang2019}. 
Research in this direction is actively growing, yet many such phenomena remain to be discovered.

This paper discusses the prediction of the accuracy of trained neural networks, using \emph{only their weights} as inputs. 
Specifically, we consider convolutional neural networks (CNNs) trained on standard datasets for the popular task of image classification.
We see this study as a step towards gaining a deeper understanding of neural network training and performance.
Understanding what can be said by looking at the trained weights can be useful in understanding the training process in general.
It can also have practical applications such as early stopping of unsuccessful training runs \cite{Domhan2015}. 

As a first step in this direction we study CNNs trained in the \emph{under-parameterized} regime, in which the observed train and test accuracies do not differ substantially. 
Then we show that our findings appear to transfer to the \emph{over-parameterized} regime \cite{belkin2018}. 
We demonstrate (Section~\ref{sec:to_demogen}) that the predictor trained on a collection of very small CNNs is capable of ranking large ResNet models according to train/test accuracy fairly well by looking only at the ResNet's weights.

The studies presented in this paper may raise more questions than they answer, but we hope that this will serve as starting point for other researchers to make progress in understanding deep learning phenomena. 
The main contributions of this paper are:\\[-2em]
\begin{itemize}
\addtolength{\itemsep}{-.5ex}
\item We propose a new formal setting that captures the approach and relates to previous works.
\item We introduce a new, large dataset with strong baselines and discuss extensive empirical results. The data is of a new modality, mapping trained weights of neural networks to their accuracy.
\item The experiments show that, somewhat surprisingly, it is possible to predict the accuracy using trained weights alone. Furthermore, only few statistics of the weights are sufficient for high accuracy in prediction.
\item Experiments on transfer of prediction across architectures and datasets show that it is possible to rank neural network models trained on an unknown dataset just by observing the trained weights, without ever having access to the dataset itself.
\end{itemize}

Next, we describe a formal setting that considers this and related tasks (Section~\ref{sec:Setting}) and discuss related work (Section~\ref{sec:relwork}). 
We introduce a new dataset for this task and present empirical results on our dataset (Section~\ref{sec:SmallCNNZoo}). 
We also discuss the performance of the resulting predictors under domain shift (Section~\ref{section:transfer}).

\section{Formal setting}
\label{sec:Setting}
Consider a fixed unknown data-generating distribution $\P(X, Y)$ defined over $\X\times \Y$, where $\X$ and $\Y$ are input and output domains, respectively.
In the context of this paper, $\X$ will be the space of images and $\Y$ will be a set of class labels.
We observe a \emph{training set} of input-output pairs $S_N := \{(X_i, Y_i)\}_{i=1}^N$ sampled i.i.d.\ from~$\P$.

We will train CNNs on $S_N$ using hyperparameters $\lambda$ and get a particular weight vector $W = \A(S_N, \lambda)$, where $\A$ denotes the learning procedure and $W$ may be considered a flattened vector containing all the weights.
The hyperparameters $\lambda$ include architecture-specific details (e.g.\:number of layers and activation function), optimizer-specific details (e.g.\:learning rate and initialization variance), and other parameters (e.g.\:weight regularization and fraction of the training set to use). 
Notice that the training method $\A$ may have internal sources of stochasticity, including order of examples in mini-batches or weight initialization.
Also note that depending on $\lambda$, the weight vector $W$ may be of a variable dimension (e.g.\ for varying number of layers). 

We will denote the function realized by the CNN with weights $W$ using $h(\cdot\,; W)\colon \X \to \Y$.   
This function has the \emph{training accuracy} $\frac{1}{N} \textstyle\sum_{i=1}^N \mathbbm{1}\{h(X_i; W) = Y_i\}$ and the \emph{expected accuracy} $\mathbb{E}_{(X, Y) \sim \P}\bigl[\mathbbm{1}\{h(X; W) = Y\}\bigr]$ denoted with $\widehat{\Acc}(W, S_N)$
and $\Acc_{\P}(W)$, respectively.

\begin{figure}[tb]
\centering
\vskip 0.01in
\begin{tikzpicture}[force/.style={},
    ellips/.style={ellipse, minimum width=1.0cm,
    align=center,fill=black!10,minimum 
    height=1.0cm,>=stealth'}
]

\node [ellips](Config) {$\lambda$}; 
\node [ellips, force, right=1.5cm of Config] (Weights) {$W$}; 
\node [ellips, force, right=1.5cm of Weights] (Accuracy) {$\Acc_{\P}(W)$};

\path[->,thick]   
     (Config) edge node[above, midway] {$S_N$}  (Weights)
     (Config) edge [bend right=20, draw=red, style=dotted] (Accuracy)  
     (Weights) edge[draw=blue] node[above, midway] {$\P$} (Accuracy);
\end{tikzpicture}
\vskip-1ex
\caption{Diagram of the learning setting. 
Nodes contain hyperparameters $\lambda$, CNN weights $W$, and expected accuracy $\Acc_{\P}(W)$.
Edges are labeled with the information necessary for the mapping: the training dataset $S_N$ and the data-generating distribution $\P$.}
\label{fig:diagram}
\vskip -.05in
\end{figure}
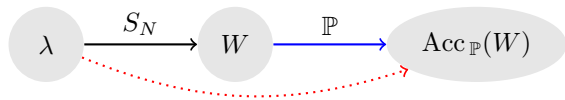

The goal discussed in this paper is to predict a CNN's expected accuracy by looking at its weights~$W$.
Importantly, since the data distribution $\P(X, Y)$ is \emph{fixed}, the mapping $W \mapsto \Acc_{\P}(W)$ that we want to learn (blue arrow in Figure \ref{fig:diagram}) exists and is defined uniquely. 
Unfortunately, it is unknown to us, as well as $\P$, and to this end we need to estimate it with a predictor $\hat{F} \colon \W \to [0, 1]$.

To build an estimator $\hat{F}$ we need to specify how to measure its quality. In other words, we need to measure how similar the mappings $\Acc_{\P}(\cdot)$ and $\hat{F}(\cdot)$, both defined on~$\W$, are.
Since this work is motivated by studying CNN training, we will not compare the two on the \emph{entire space} $\W$ but rather focus on the subset consisting of weights that can be actually obtained as a result of training.
We propose to generate a set of hyperparameter configurations $\lambda_1,\dots,\lambda_K$ 
and then train $K$ different CNNs $W_k = \A(S_N; \lambda_k)$ 
on the training set $S_N$.
We cannot compute the exact values of $\Acc_{\P}(W_k)$, but we can estimate them well
using the \emph{test accuracy} $T_k := \widehat\Acc(W_k, S_M')$ measured on the separate \emph{test set} of i.i.d.\:input-output pairs $S'_M:=\{(X'_j, Y'_j)\}_{j=1}^M$ sampled from $\P$ independently of $S_N$.
Finally, we can train the estimator $\hat{F}$ by minimizing its Mean Squared Error (MSE) on the \emph{CNN collection} $\C:=\{(W_k, T_k)\}_{k=1}^K$. 

{\bf Why use only weights?}
The framework proposed above already makes use of the dataset $S_N$ by training CNNs $W_1,\dots,W_K$ on it.
This means that the estimator $\hat{F}$ and, as a consequence, its predictions, \emph{implicitly} depend on $S_N$.
A~natural idea would be to make the dependence on $S_N$ more explicit: e.g.\ by holding out some part $S\subseteq S_N$ and returning $\widehat{\Acc}(W, S)$ as a prediction for the accuracy of the CNN~$W$.
Based on decades of theoretical and practical ML experience, this approach will likely provide a very strong baseline for the task of predicting the accuracy.
So why are we considering predictors $\hat{F}$ that \emph{only look at weights} and not utilize $S_N$ explicitly?

The main reason is that predicting the accuracy is only an indirect goal of this study.
Ultimately we hope to gain insights about DNN training and generalization by understanding the structure of network weights, which are some of the most prominent characteristics of the DNN.
Other minor advantages of not choosing another set $S\subseteq S_N$ to compute $\widehat{\Acc}(W, S)$ can be of a more practical nature: 
supporting prediction with less computational effort than an inference pass over $S$ requires.

\vspace{0.1ex}
\subsection{Predicting from hyperparameters}
\label{section:from-hyper}
Another important and related question is to what extent the test accuracy of $W = \A(S_N, \lambda)$ can be predicted from the hyperparameters $\lambda$ that were used to train it.
Once we fix the training set $S_N$ and the \emph{random seed}, which determines the learning procedure's internal source of stochasticity, 
there exists a unique deterministic mapping $\lambda\mapsto \Acc_{\P}(W)$ (dotted red arrow in Figure \ref{fig:diagram}) and we may try to estimate it using the same scheme as described above.
While the Bayes error of both using $\lambda$ or the resulting weights $W$ for predicting the accuracy is 0, in practice the two problems may have different sample complexities.

If the training set $S_N$ and/or the random seed are not fixed but instead generated each time we train the CNN, the mappings $\lambda \mapsto \A(S_N, \lambda)$ and, as a consequence, $\lambda \mapsto \Acc_{\P}(W)$ both become \emph{stochastic}.
In this case the estimation is possible only up to the noise introduced by the variance of $S_N$ and/or different choices of the seed. 

\vspace{0.1ex}
\subsection{Domain shift}
\label{section:domain-shift}
Does the learned estimator $\hat{F}$ generalize to yet unseen data distributions $\P$ or hyperparameter configurations $\lambda$? In other words, if we were to train an estimator~$\hat{F}$ on CNNs which were themselves trained on CIFAR10, how accurately would $\hat{F}$ predict the test accuracy of a CNN trained on SVHN?
We will refer to this setting as \emph{domain shift}.
A~priori, even if we solve the original problem well on CIFAR10, there are no guarantees that the estimator would perform well for SVHN.
The same applies to a change in the architecture.

Rather than discovering properties of DNNs that are specific to a particular dataset or architecture (which nevertheless could be interesting on its own), we are even more interested in those that hold \emph{across various datasets and architectures}.
In that sense, domain shift provides a setting close to what we actually are interested in: 
observing any sort of positive transfer between different datasets and architectures would indicate that there are properties of DNNs that transfer.
Our goal is to demonstrate the existence of these invariant properties and study them.

\section{Related work}
\label{sec:relwork}

There are only few works that consider the problem setting described in Section \ref{sec:Setting}. 
The most relevant of these are \cite{Jiang19,Yak19,eilertsen2020,martin2020,martin2020b}.

The overall setting and motivations of \citet{eilertsen2020} are similar to ours. 
However, the main difference is that instead of predicting the accuracy, the authors focus on predicting the hyperparameters $\lambda$ using the weights $W$ (the opposite direction of the black arrow in Figure \ref{fig:diagram}).

Concurrent works from \citet{martin2020, martin2020b} confirm our findings by showing that more complex statistics derived from weight matrices \cite{martin2018} correlate well with the performance of state-of-the-art models in vision and language processing.

\citet{Jiang19} and \citet{Yak19} both  investigate how to predict the generalization gap, i.e.\ the difference between training and test set performance, of a neural network based on the hidden activations of training set examples.
\citet{Jiang19} train large CNN/ResNet architectures on CIFAR datasets and approximate the minimal distances to the class boundary for each data point in each hidden layer. 
They use this \emph{margin distribution} to train a linear regressor 
that predicts generalization gaps. 
\citet{Yak19} expand upon this work by training a large number of small fully-connected networks on different variations of a generated 
{spiral} dataset. 
They replace the linear predictor with a recurrent neural network to handle varying neural network depth, and show that predictions transfer between small fully-connected architectures and varying synthetic datasets.
Both works heavily rely on the margins in the intermediate layers of the networks.
These margins can not be computed analytically and require a computationally expensive approximation procedure \cite{Elsayed2018}, which is not guaranteed to be accurate.
Margin approximation also involves an inference pass over the training set~$S_N$.
Our estimators $\hat{F}$ use \emph{only weights} of the networks (or their simple statistics) to predict the accuracy. 
As the weights are (one of) the most important characteristic of a trained DNN, it is interesting to study this connection without requiring information about the training set $S_N$.
We show that these estimators transfer to networks trained on unobserved natural image datasets and with ResNet32 architectures.
Finally, experimental design utilized in these previous works may lead to an undesirable \emph{leakage}, as discussed in Section~\ref{section:the-dataset}.

\citet{DeChant19} train ResNets and other large architectures on CIFAR and ImageNet datasets. 
They demonstrate that it is possible to tell whether or not the network will make a mistake on \emph{one particular image} by looking at the activations of that image in the network's layers.

The relation between the train and test accuracies is the central question of statistical learning theory \cite{Vapnik98, Shalev2014}.
\citet{Jiang20} recently performed a large scale empirical study analyzing correlation between various generalization error bounds and network performance.

A problem somewhat similar to ours has been studied in the context of hyperparameter optimization and neural architecture search (NAS).
\cite{Streeter19,Streeter19icml} propose procedures that select good hyperparameter values based on previous exploration.
To apply early stopping to unsuccessful runs, \citet{Swersky2014} and \citet{Domhan2015} predict the final performance of a neural network based on few training iterations. 
Similar techniques were applied in NAS to select candidate architectures, where the prediction is usually based on hyperparameters, architectures, information about the dataset, and performance measurements of similar architectures, 
see \cite{Baker2017,Istrate2019} and references therein.

\section{Experiments: Small CNN Zoo}
\label{sec:SmallCNNZoo}
Results reported in this section are based on a new dataset which we call the \emph{Small CNN Zoo}\footnote{Made publicly available together with the code reproducing the experiments at \url{https://github.com/google-research/google-research/tree/master/dnn_predict_accuracy}}.
It contains weights of a \emph{fixed CNN architecture} trained on 4 different image datasets using a large number of different hyperparameter configurations. 
For each network, accuracy and cross-entropy loss on the train and test data are available.

\subsection{The Small CNN Zoo dataset}
\label{section:the-dataset}
To enable predicting accuracy from the flattened weight vector, we keep the number of weights in the architecture small: 3 convolutional layers with 16 filters each, followed by global average pooling and a fully connected layer, for a total of 4\,970 learnable weights. 
As a result, the best test accuracies we obtain on CIFAR10 and SVHN are 56\% and 78\%, respectively, which is far below state of the art. However, it is worth pointing out that the smallest CNN architectures achieving above 90\% test accuracy on CIFAR10 that we are aware of require on the order of $10^6$ parameters \cite{Lin2014, Springenberg2015}, i.e.\ 200x more, and work on RGB inputs, while we ignore away color information.

We train on 4 natural image classification problems: 
MNIST \cite{lecun2010mnist}, Fashion MNIST \cite{Xiao2017}, grayscale CIFAR10 (CIFAR10-GS) \cite{Krizhevsky09learningmultiple}, and grayscale SVHN (SVHN-GS) \cite{Netzer2011}.
Global average pooling and using grayscale allows us to apply the same architecture across all four datasets.

For each dataset, we sample 30k different hyperparameter configurations chosen independently at random from pre-specified ranges (listed in the Supplemental \ref{appendix:base-cnn-sweep}). We vary optimizer, learning rate, type of initialization and its variance, fraction of the training examples to use, activation function, dropout rate, and $\ell_2$-regularization of weights.
We use one random seed per hyperparameter configuration.
We did not use data augmentation or batch normalization.

Instead of stopping training when networks converge or reach a certain level of accuracy, we train each CNN for 86 epochs.
We do so because we want to study CNNs under general conditions: properties discovered by only looking at converged models may not hold for intermediate steps.

Finally, we discard the models in which numerical instabilities (e.g.\ infinite gradients) were detected.
This process leads to 4 CNN collections: 
$\CM$ with 29\,996 CNNs for MNIST, 
$\CF$ with 29\,999 for Fashion MNIST, 
$\CC$ with 29\,999 for CIFAR10-GS, 
and $\CS$ with 29\,987 for SVHN-GS.
The Small CNN Zoo is the union 
of these 4 collections. 

\begin{figure}[tb]
\begin{center}
\centerline{\includegraphics[width=\columnwidth]{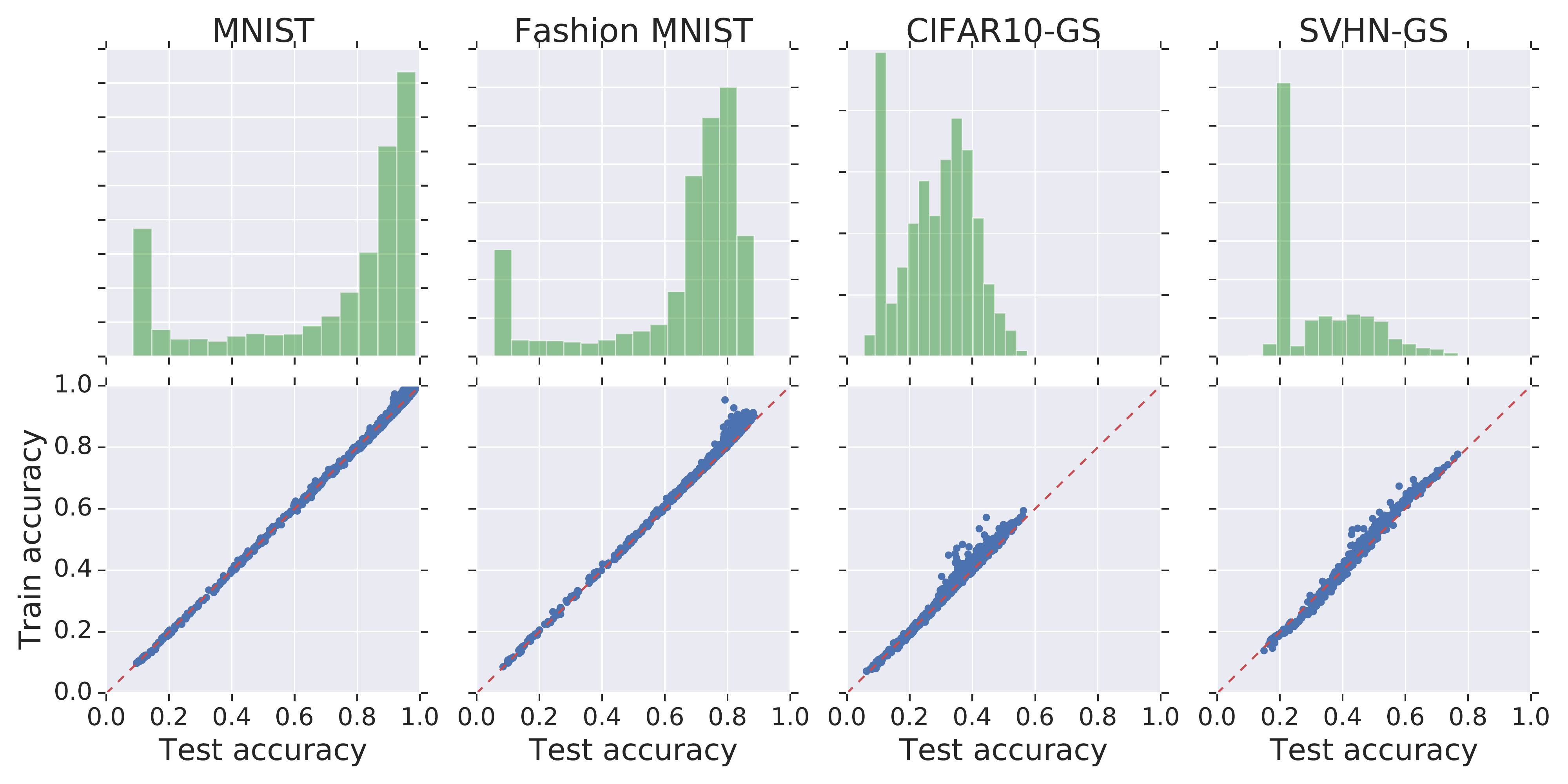}}
\caption{Distribution of the networks from the Small CNN Zoo collection over their test accuracy (first row) and their training/test accuracies (second row).}
\label{fig:test_acc_hist}
\end{center}
\vskip -0.2in
\end{figure}

The distribution of the CNN models with respect to their test/train accuracy is reported in Figure~\ref{fig:test_acc_hist}.
MNIST, Fashion MNIST, and CIFAR10-GS all have balanced classes and the histograms peak at around 10\%---the accuracy of a random or constant prediction.
SVHN-GS is unbalanced with the largest class containing around 19\% of the samples.
Here many models seem to converge to the constant majority class prediction, which explains the shifted peak.

We do not observe overfitting in the Small CNN Zoo dataset, even though some of the models were trained only on 10\% of the training examples.
Likely, this is due to the small architecture used.
Based on this dataset we may gain insights on why and how neural networks \emph{train}, but it is less likely that the dataset will directly lead to deeper understanding \emph{generalization}.

{\bf Why not use multiple seeds?}
We use one random seed per hyperparameter configuration. 
This avoids having models that are too similar between the train and test splits of the CNN collections, which possibly leads to a \emph{leakage}.
In~fact, this may point to a possible shortcut taking place in the studies of \citet{Jiang19}.
The authors used 3 random seeds per hyperparameter configuration and did not enforce that the models trained with the same hyperparameters (but different random seeds) were allocated to the same split.
Inspecting their dataset closer shows that the variance in generalization gap between the networks that only differ in random seed is orders of magnitude smaller than the average variance between all networks ($10^{-5}$ vs $10^{-3}$).
A similar shortcut may take place in the studies of \citet{Yak19}.
Here, the authors did not use the same hyperparameters with different seeds for training, but they trained networks with the same hyperparameters on versions of the synthetic datasets that were generated using different seeds.

\subsection{Training the estimators}
\label{section:training-estimators}
Once we have the CNN collection $\C:=\{(W_k, T_k)\}_{k=1}^K$ with weights $W_k\in\R^{4\,970}$ and their test accuracies $T_k$, we can start training various estimators $\hat{F}\colon W\to[0, 1]$.

{\bf Types of estimators}
We explore three different estimators: logit-linear model (L-Linear), gradient boosting machine using regression trees (GBM), and a fully-connected DNN.
All three methods were trained to minimize MSE.
Each of these 3 methods comes with its own hyperparameters and initial experiments showed that it is important to tune them.

For the logit-linear model, we train weights and offsets using 
mini-batch SGD/Adam varying the learning rate, batch size, initialization, and $\ell_2$-regularization.
We use LightGBM \cite{Ke2017} to train the GBM model and vary the number of leaves and maximum depth of the trees, the learning rate, $\ell_1$ and $\ell_2$ regularization, and parameters for the features/examples subsampling.
We use a feed-forward fully-connected architecture for the DNN model with ReLU activations and sigmoid transform.
We train it with mini-batch SGD/Adam varying the learning rate, number of layers and their width, $\ell_2$-regularization, initialization type and variance, and batch size.

{\bf Input features}
We investigate several ways of preprocessing the weight vectors $W$ before feeding them to the estimators:
(1) Using flattened parameters (weights/kernels and biases) of a single $\ell$-th layer $W^{\ell}$, $\ell=1,\dots,4$ ($W^4$ stands for the last fully connected layer);
(2) Using statistics $\widetilde{W}$ of the entire flattened vector consisting of 7 real numbers: the mean, the variance, and $q$-th percentiles for $q\in\{0, 25, 50, 75, 100\}$;
(3) Computing the above statistics for each layer $\ell=1,\dots,4$ separately, while processing kernels and biases independently, and concatenating the results, which yields $4 \times 2 \times 7 = 56$ real-valued features $\wstats$;
(4) Computing $\ell_1$ or $\ell_2$ norms for each layer $\ell=1,\dots,4$ separately, while processing kernels and biases independently, and then concatenating the results, which yields $4 \times 2 = 8$ real-valued features $W_{\!L}^{\ell_1}$ and $W_{\!L}^{\ell_2}$.

{\bf Training protocol and metrics}
Each of the 4 CNN collections is divided into two splits: 15k CNNs are used for the \emph{training split} and the remaining ones were held out for the \emph{test split}. 
The entire training and hyperparameter selection for the models took place on the training splits.
The test splits are used \emph{only once} to evaluate the single best model that we chose based on the 3-fold cross-validation performed on the training split.

We performed hyperparameter selection by evaluating 1k unique hyperparameter configurations sampled randomly and independently from pre-specified ranges 
for every combination of estimator type, input features, and CNN collection.

In all experiments we use MSE as the training objective. We also compute the mean absolute deviation and the \emph{coefficient of determination} or $R^2$ score.
The $R^2$~score normalizes the MSE of the estimator $\hat{F}$ by the MSE of the \emph{best constant} prediction.
Larger $R^2$ scores correspond to better predictions and the score never exceeds 1.
For further details on the Small CNN Zoo dataset and the experimental setup we refer to Supplementary \ref{appendix:small-cnn-zoo-specs}.

\begin{table}[tb]
\vskip -0.1in
\caption{$R^2$ scores for predicting test accuracies of CNNs trained on CIFAR10-GS with different input features ({columns}) and different estimators ({rows}).
GBM is on par or better than DNN, and significantly better than L-Linear.
All std.\,dev.\ (w.\,r.\,t.\:training models on three different folds of the cross-validation) for numbers in this table were below $0.005$.
}
\label{table:SCNNZ-methods}
\begin{center}
\begin{small}
\begin{sc}
\def\arraystretch{1.5}
\addtolength\tabcolsep{.5ex}
\begin{tabular}{ccccc}
\hline
  & ${W}^4$ & ${W}$ & $\widetilde{W}$ & $\wstats$ \\ \hline
 L-Linear & $0.707$ & $0.662$ & $0.186$ & $0.727$ \\
 GBM & $0.969$ & $0.970$ & $0.914$ & $0.984$ \\
 DNN & $0.968$ & $0.954$ & $0.897$ & $0.980$ \\ \hline
\end{tabular}
\end{sc}
\end{small}
\end{center}
\vskip -0.2in
\end{table}

\subsection{Empirical results}
\label{section-main-results}
In the experiments, GBM and DNN models always produce significantly better results than the logit-linear model.
In some cases, the DNN model achieves slightly better results than GBM, but overall it is on par or significantly worse than GBM.
These conclusions hold across all 4 datasets
and the corresponding results are shown for one of the datasets (CIFAR10-GS) and a selection of input features in Table \ref{table:SCNNZ-methods}.
In the interest of space we therefore only report the results for GBM in the following.
All numbers for other models can be found in Supplementary \ref{appendix:predictors-detailed}.

Table \ref{table:SCNNZ-gbm} presents the results of training the GBM models with different input features on the 4 CNN collections. 

{\bf Using flattened weights}
First, we notice that a naive baseline of using the entire flattened vector $W$ already achieves a rather strong performance across all 4 datasets.
Interestingly, almost the same performance can be recovered just by using the parameters of the last dense layer $W^4$, while using any other (convolutional) layer alone results in a noticeably worse performance. 
This observation is consistent with \emph{feature importance} measurements produced by the GBM model (Supplementary \ref{appendix:gbm-importances}), which indicate that parameters of the last dense layer were among the most informative and frequently used ones.

{\bf Using weight statistics}
Results based on the per-layer statistics of the weights $\wstats$ are the best obtained across all 4 CNN collections.
In particular, they are significantly better than results based on the entire weight vector $W$.
At~first glance this may look surprising, because $W$ contains sufficient information to recover $\wstats$.
However, computing quantiles requires sorting numbers and presumably neither the GBM nor DNN estimators have capacity to do this.
Also, compared to the entire weight vector $W\in\R^{4\,970}$ or the weights of the last dense layer $W^4 \in \R^{170}$ the feature vector of statistics $\wstats \in \R^{56}$ is relatively low-dimensional.
This may provide an additional explanation of superior performance of $\wstats$: sample complexity of the regression problem is generally known to grow with the dimension of the feature space \cite{tsybakov2009}.

Notably, \citet{eilertsen2020} also report a strong performance of the per-layer statistics in their work.

{\bf Using weight norms}
We also tried using the $\ell_1$ and $\ell_2$ norms of the weights as features $W^{\ell_1}_L$, $W^{\ell_2}_L$.
Norms traditionally play an important role in the statistical learning theory \cite{Neyshabur2015, Bartlett2017} and are still actively used in practice to regularize DNNs with weight decay.
In contrast to weight decay, which is commonly implemented by adding \emph{the sum of the norms across all layers} multiplied by a single regularization coefficient to the objective, we kept the norms for different layers separate.
This should provide more flexibility for the estimator.
Table \ref{table:SCNNZ-gbm} (first block) shows that the estimators based on the norms perform slightly (but statistically significantly) worse than the ones using weight vectors $W$ or $W^4$.

{\bf Interpreting the $R^2$ score and MSE values}
MSE provides an \emph{absolute measure} of the model performance and on its own does not tell us much about the model: the value of $10^{-4}$ can correspond to a good and bad performance depending on the problem.
The $R^2$ score is a \emph{relative measure}: it compares the MSE of the model to the MSE of a constant prediction.
Moreover, $R^2$ score is scale invariant and multiplying the outputs by a constant won't change the metric.
In Table \ref{table:SCNNZ-gbm} we use the
$R^2$ scores because we find them slightly easier to interpret: a non-positive value indicate that we are not doing better than fitting a constant predictor and values close to 1 point at stronger performance.
The MSE values are reported in the Supplementary \ref{appendix:predictors-detailed}
and scatter plots with raw predictions and true targets can be found in Supplementary \ref{appendix:transfer-plots}.

\begin{table}[tb]
\vskip -.1in
\caption{$R^2$ scores for predicting test accuracies of CNNs trained on various datasets ({columns}) with GBM using different input features ({rows}).
Best numbers for each dataset are in boldface.
The~largest std.\,dev.\ (w.\,r.\,t.\:training models on three different folds of the cross-validation) across all numbers in this table was $0.002$.
See Sections \ref{section-main-results} and \ref{section:ablation} for row descriptions.
}
\label{table:SCNNZ-gbm}
\begin{center}
\begin{small}
\begin{sc}
\def\arraystretch{1.3}
\vskip -0.05in
\begin{tabular}{lcccc}
\hline
& \multicolumn{1}{p{.16\linewidth}}{\centering MNIST}
& \multicolumn{1}{p{.16\linewidth}}{\centering Fashion\\ MNIST}
& \multicolumn{1}{p{.16\linewidth}}{\centering CIFAR10\\-GS}
& \multicolumn{1}{p{.16\linewidth}}{\centering SVHN\\-GS}\\
\hline
${W^1}$              & 0.977 & 0.982 & 0.959 & 0.936 \\
${W^2}$              & 0.966 & 0.975 & 0.926 & 0.895 \\
${W^3}$              & 0.969 & 0.973 & 0.928 & 0.900 \\
${W^4}$              & 0.987 & 0.989 & 0.969 & 0.967 \\
${W}$                & 0.988 & 0.989 & 0.970 & 0.971 \\
${W}_{\!L}^{\ell_1}$     & 0.983 & 0.982 & 0.960 & 0.967 \\
${W}_{\!L}^{\ell_2}$     & 0.981 & 0.983 & 0.960 & 0.971 \\
$\widetilde{W}$    & 0.953 & 0.955 & 0.914 & 0.908 \\
$\wstats$& {\bf 0.993} & {\bf 0.993} & {\bf 0.984} & {\bf 0.986} \\
\hline
{$\lambda$}                     & 0.918 & 0.924 & 0.934 & 0.935 \\
${\lambda}_{\mathrm{LR}}$        & 0.024 & 0.035 & 0.015 & 0.034 \\
${\lambda},\!{W}$         & 0.990 & 0.991 & 0.979 & 0.978 \\
$\widetilde{{W}}_{\!L}^4$           & 0.973 & 0.976 & 0.941 & 0.946 \\
$\widetilde{{W}}_{\!L}^{1,4}$         & 0.989 & 0.989 & 0.971 & 0.971 \\
\hline
\end{tabular}
\end{sc}
\end{small}
\end{center}
\vskip -0.15in
\end{table}

\subsection{Ablation studies}
\label{section:ablation}
Table~\ref{table:SCNNZ-gbm} (upper block) shows that the parameter vector of a trained CNN alone contains a strong signal regarding the network's accuracy.
To understand more about the nature of this signal, we performed additional studies.

We tried several other input features for the estimators, including
(i) the hyperparameter configuration $\lambda$ (containing 7 parameters) used while training the CNN, 
(ii) the concatenation $(\lambda, W)$ of the hyperparameters $\lambda$ with the entire weight vector $W$, and
(iii) the weight statistics similar to $\wstats$ computed only for a subset of the layers: $\widetilde{W}_{\!L}^4$ for the final dense layer and $\widetilde{W}_{\!L}^{1,4}$ for the combination of the first convolutional and the final dense layers.
The results are reported in Table \ref{table:SCNNZ-gbm} (second block).

{\bf Hyperparameters} 
As discussed in Section \ref{section:from-hyper}, the Bayes error of the predictor based on either hyperparameters $\lambda$ or weight vectors~$W$ is 0, but sample efficiency may differ.
The results show that for the Small CNN Zoo predicting the accuracies with weights is easier than with hyperparameters.
We also tried predicting from individual hyperparameters 
to see if there was a single parameter sufficient to recover the signal.
They all gave similar bad results.
For the reference, we include the results for the~\emph{learning rate}~${\lambda}_{\mathrm{LR}}$.
Predicting with both $\lambda$ and weights $W$ does not improve on predicting with $W$ alone.

{\bf Statistics for subsets of layers}
Motivated by the fact that using the weights of the last dense layer $W^4$ is as good as using the whole weight vector $W$ we tested whether statistics for a subset of the layers is enough to recover the performance based on $\wstats$.
Curiously, the statistics of the last dense layer $\widetilde{{W}}_{\!L}^4$ perform worse.
The results improve if we add the statistics of the first convolutional layer $\widetilde{{W}}_{\!L}^{1,4}$, but they are still slightly worse than with all layers.

{\bf Permutation and scale invariance}
We also examined how the estimator's predictions change as we modify its inputs.
Notice that two ReLU CNNs with parameters $W$ and $c\cdot W$ have exactly the same test/train accuracy (but not the same cross-entropy loss) for any real value $c>0$, because their outputs $h(X; W)$ and $h(X; c\cdot W)$ coincide for all inputs~$X$.
The same is true for any CNN if we permute the order of filters/channels consistently across all layers.
We want to emphasize that we did not incorporate these inductive biases in any of the estimators we trained.
Nevertheless, it may be interesting to test whether these (or similar) invariances emerge naturally in the trained estimators.

For a given estimator $\hat{F}$ trained with the entire weight vectors $W$, we tested several ways of modifying its inputs $W\mapsto \varphi(W)$, including multiplying it with various positive factors and permuting it in several different ways.
Then we looked at the absolute difference $|\hat{F}\bigl(\varphi(W)\bigr) - \hat{F}(W)|$ across multiple CNNs $W$ (from the test split of the same CNN collection $\hat{F}$ was trained on) and various types of modifications $\varphi$.
We report a short summary of this study here.
Details can be found in Supplementary \ref{appendix:invariance-study}.

The Mean Absolute Deviation (MAD) of modifications $\varphi$ that we tried spanned a range between $0.01$ and $0.13$.
Scaling the weights $\varphi(W) = c\cdot W$ with $c\in\{2, 10, 100\}$
or
permuting parameters within each of the first 3 convolutional layers leads to MADs less than $0.05$.
The estimator is more sensitive to permutations within the final dense layer, which leads to a MAD of 0.06.
Global permutation of the entire vector $W$  (without preserving the layers)
or
scaling with small constants $c\in\{10^{-1}, 10^{-3}\}$
all lead to a MAD larger than $0.11$.
Summarizing, the estimator is not too sensitive to the order of parameters in the convolutional layers, and much more sensitive to permutations within the final dense layer.
The estimator is invariant to scaling the weights with positive factors larger than 1 and changes its predictions significantly for factors smaller than 1.

\subsection{Understanding observed behaviors: first steps}

Seeing that very few values extracted from the weights already lead to good predictions, it is natural to ask if these predictions can be (at least partially) reduced to simple, human-interpretable rules. 
In informal experiments we explored different approaches such as GBM feature importances, LASSO, and univariate feature selection, but did not observe any clear and consistent signals.
Thus, we manually inspect the CIFAR10-GS CNN collection $\CC$. 

\begin{figure}[tb]
\begin{center}
    \centerline{\includegraphics[width=0.49\linewidth]{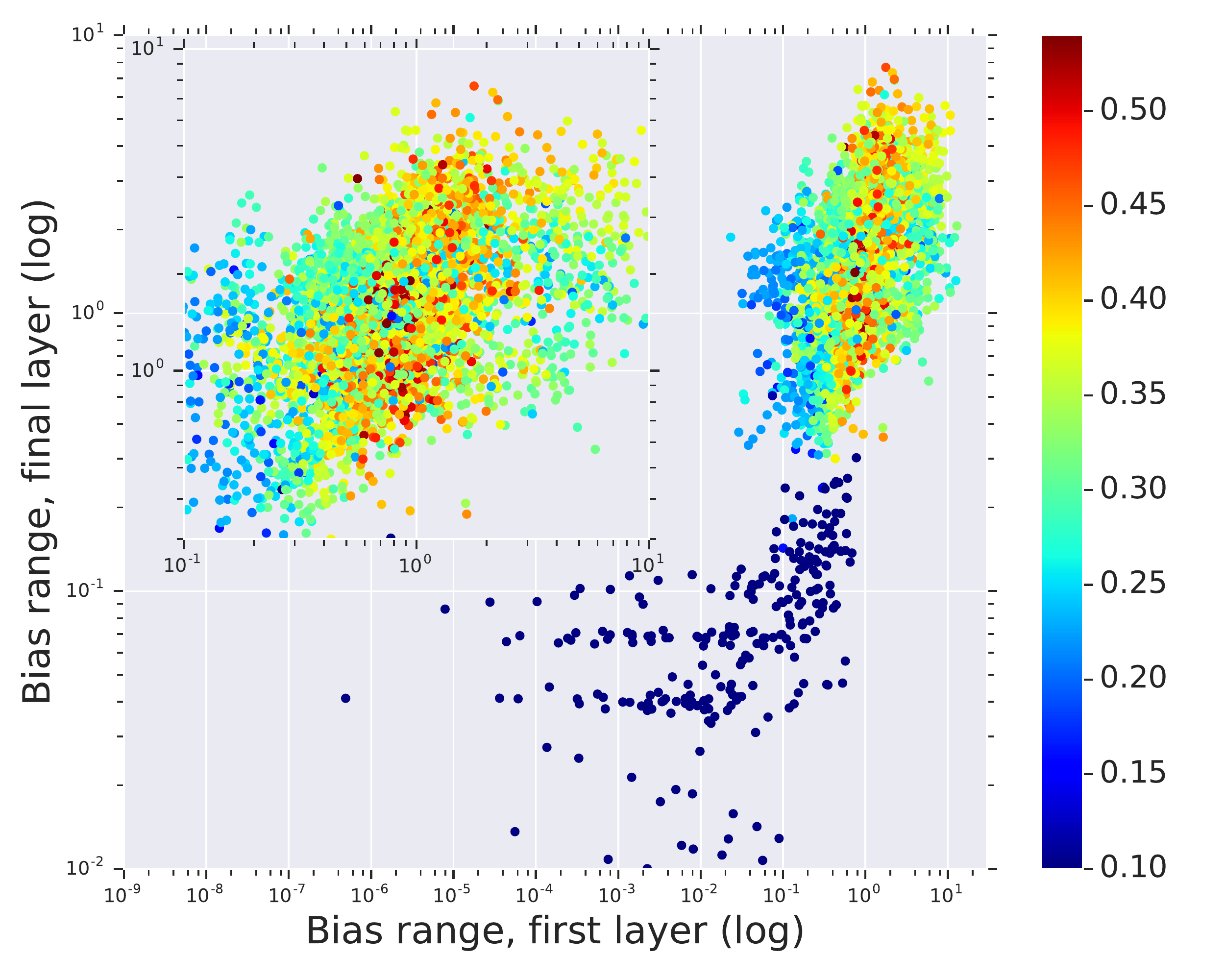}
    \includegraphics[width=0.49\linewidth]{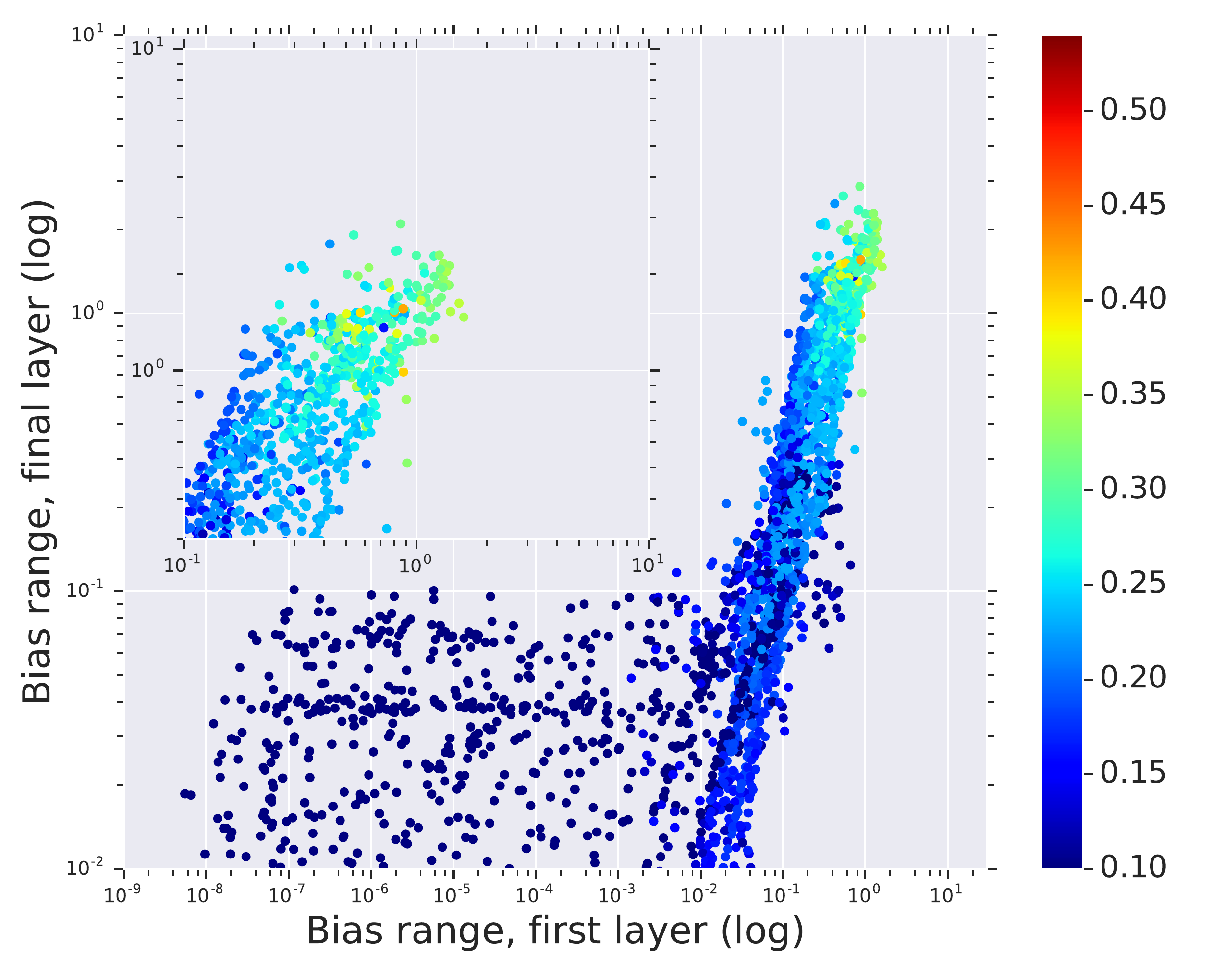}}
    \caption{Scatter plots of the networks trained on CIFAR10-GS, colored by test accuracy (best viewed in color). 
    Bias range width ($\mathrm{max}-\mathrm{min}$) in first layer (x-axis) and last layer (y-axis) together with the upper-right corners zoomed in. 
    Networks trained with Adam/RMSProp (left) and SGD (right).}
    \label{fig:scatter}
\end{center}
\vskip -.2in
\end{figure}

As discussed above, we observe that the prediction works well when only considering simple statistics of the weights as inputs for the predictor. 
When predicting based on just one full network layer, the first and last layers of the network are most useful. 
Looking manually into weight statistics we noticed that the range ($\mathrm{max}-\mathrm{min}$) of the \textit{biases} in the first and last layer is often correlated with the network's accuracy. 
(Note that biases in the Small CNN Zoo were initialized to~0.) 
We do not claim any particular significance of this observation, but it can be used to \textit{generate} hypotheses that could then be verified in further experiments.
For example, we can visualize the networks' performance in a 2D scatter plot using those two ranges (Figure~\ref{fig:scatter}). It seems surprising how well these two measurements separate the data (at least visually) already. 

We notice that networks trained on CIFAR10-GS with SGD (Figure~\ref{fig:scatter}, right) perform significantly worse than those trained with Adam/RMSProp (Figure~\ref{fig:scatter}, left).
It is also interesting to note how the majority of the networks trained with SGD align along a line in this 2D space.
Among the networks trained with Adam/RMSProp, we observe two well-separated groups: 
the strongly performing ones in the upper-right corner (also depicted in the zoomed-in subplot)  
and the ones with near-chance performance (the blue ``tentacles'' in the bottom part).
Further analysis reveals that these two groups can be perfectly separated from each other by looking at the bias \textit{maxima} in the final dense layer (not shown in the plots): the bias maxima are below 0.1 for the badly performing models (the ``tentacles'') and above 0.1 for all the rest of the networks.
In future work we would like to understand better what causes these ``symptoms'' during training and investigate ways to alleviate them.

\section{Transfer to new architectures and datasets}
\label{section:transfer}
In the previous section we showed using the Small CNN Zoo dataset that strong predictors of accuracy based on weights exist.
Next we want to explore the domain shift setting introduced in Section \ref{section:domain-shift} and study whether the predictors can handle networks trained on unobserved datasets or with different architectures.
We emphasize that throughout this section the models were not fine-tuned or adjusted to the new collections in any way.

\subsection{Networks trained on unobserved datasets}
\label{section:smallcnnzoo-transfer}
\begin{figure}[tb]
\begin{center}
\centerline{\includegraphics[width=\columnwidth]{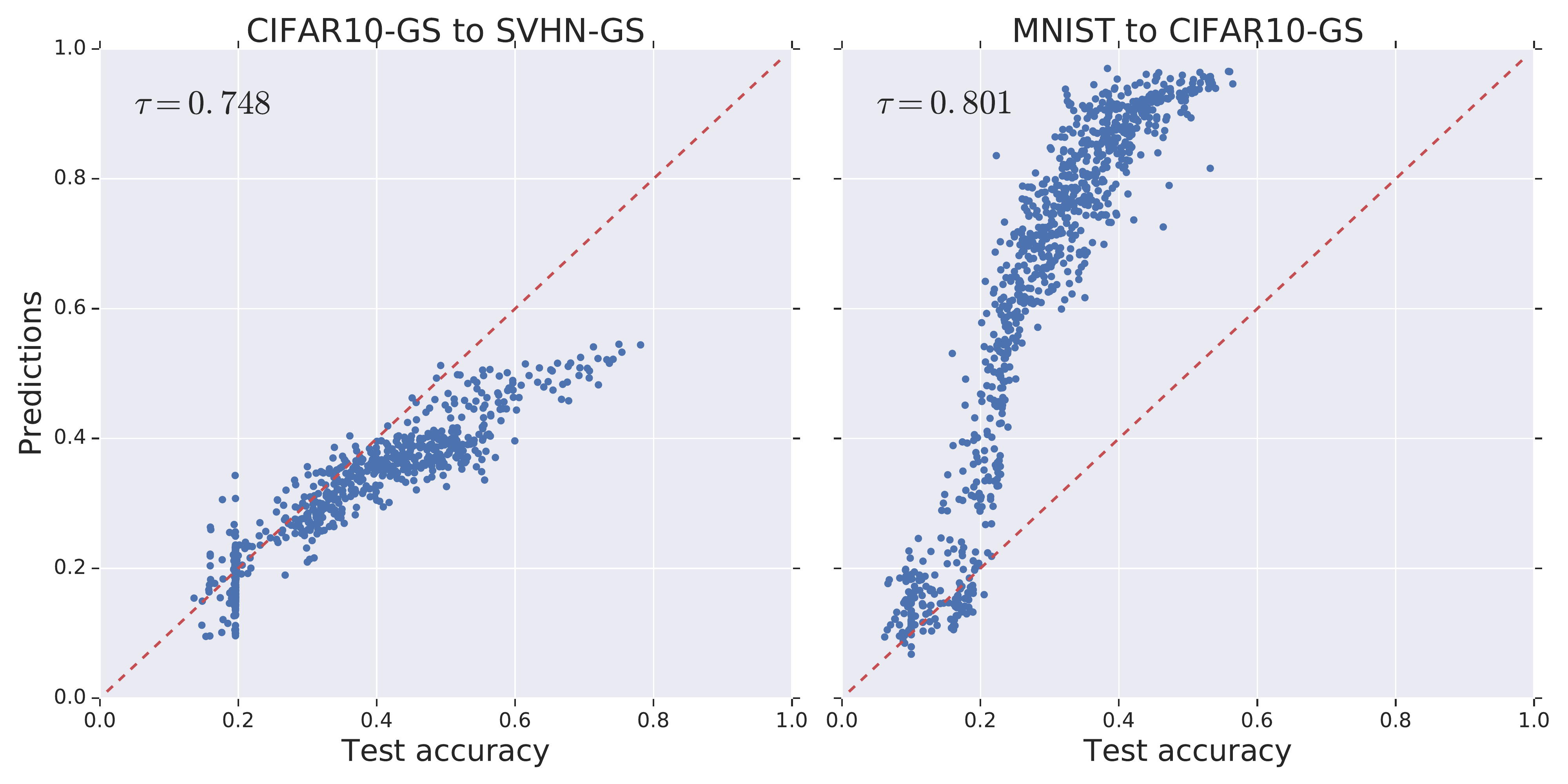}}
\caption{
Distribution of true/predicted test accuracies for networks from the SVHN-GS ({left}) and CIFAR10-GS ({right}) collections together with Kendall's $\tau$ coefficient.
Predictions were made with the GBM models trained on CIFAR10-GS ({left}) and MNIST ({right}) collections using $\wstats$.
}
\label{fig:transfer_dataset}
\end{center}
\vskip -0.3in
\end{figure}

First we look at how the GBM models transfer across the CNN collections.
Two examples of such experiments are shown in Figure \ref{fig:transfer_dataset}.
The figures demonstrate that the MSE of the predictions may not be the best metric to look at.
The~``drift'' of points away from the diagonal line (which corresponds to zero MSE) is likely due to the difference in \emph{average accuracy} between various datasets. 
Most of the networks in the MNIST collection achieve an accuracy higher than 60\%, while the best accuracy for CIFAR10-GS was 55\%.
Nevertheless, we see that networks with higher accuracy tend to receive higher prediction values.
In other words, the predictors are doing a reasonable job in \emph{ranking} the networks.
We can use Kendall's $\tau$ rank correlation coefficient to measure the quality of ranking. It ranges from -1 (anti-ranking) to 1 (perfect ranking) and takes values around 0 for random ranking.

Table~\ref{table:SCNNZ-kendall} contains the values of Kendall's $\tau$ coefficient for all possible transfer experiments performed on the Small CNN Zoo (and 2D plots similar to Figure \ref{fig:transfer_dataset} are reported in Supplementary \ref{appendix:transfer-plots}).
The smallest coefficient of 0.6 corresponds to the transfer from SVHN-GS to the MNIST collection.
It is perhaps surprising that the rank test shows such a large correlation.
We want to highlight that when training CNNs on the 4 datasets we only scale the pixel values to the $[-1,1]$ interval and do not perform any other standardization.
We would expect the moments of the pixel values for the MNIST dataset to be very different from those of SVHN-GS 
and this difference in distributions to affect the form of the filters in the convolutional layers.

\begin{table}[tb]
\vskip -.1in
\caption{Kendall's rank correlation between GBM model's predictions and true test accuracies. 
The GBM model trained using layer statistics $\wstats$ as inputs on one CNN collection ({rows}) was used to make the predictions on the other ({columns}).
}
\label{table:SCNNZ-kendall}
\begin{center}
\begin{small}
\begin{sc}
\def\arraystretch{1.3}
\begin{tabular}{p{1.9cm}cccc}
\hline
& \multicolumn{1}{p{1.05cm}}{\centering MNIST} 
& \multicolumn{1}{p{1.05cm}}{\centering Fashion\\ MNIST} 
& \multicolumn{1}{p{1.05cm}}{\centering CIFAR10\\-GS} 
& \multicolumn{1}{p{1.05cm}}{\centering SVHN\\-GS}\\
\hline
MNIST       & 0.92 & 0.77 & 0.80 & 0.73\\
Fash.\,MNIST   & 0.70 & 0.92 & 0.77 & 0.65\\
CIFAR10-GS  & 0.68 & 0.68 & 0.93 & 0.75\\
SVHN-GS   & 0.60 & 0.63 & 0.74 & 0.85\\
\hline
\end{tabular}
\end{sc}
\end{small}
\end{center}
\vskip -0.1in
\end{table}

\subsection{Networks trained with different architecture}
\label{sec:to_demogen}
In this section we want to test whether predictors trained on the Small CNN Zoo can rank larger, \emph{over-parametrized} networks, capable of overfitting.
For this purpose we will use the DEMOGEN collection \cite{Jiang19}, which contains 216 Wide-ResNet32 models \cite{He2016} trained on the original (colored) CIFAR10 dataset with the best models achieving 100\% training and 93\% test accuracy.
The collection contains 72 networks for each of the three different architectures: ResNet32x1, ResNet32x2, and ResNet32x4, which differ in the number of filters.

In Section~\ref{section:ablation} we discovered that weight statistics $\widetilde{{W}}_{\!L}^{4}$ computed for the final layer of the CNN provide a strong signal for predicting a network's test accuracy.
Using these features on the CIFAR10-GS collection, GBM achieves performance that is very close to the best model overall (Table~\ref{table:SCNNZ-gbm}). Because the vector $\widetilde{{W}}_{\!L}^{4}\in\R^{14}$ has the same dimension for CNNs of \emph{any architecture}, 
we can use estimators trained on the Small CNN Zoo to make predictions for the ResNet32 models from the DEMOGEN collection.

Table~\ref{table:demogen} reports the $\tau$ coefficients demonstrating how well predictions of the GBM model trained on CIFAR10-GS CNN collection correlate with actual accuracies of the networks from DEMOGEN.
We compare to both \emph{train and test} accuracies, because, as discussed in Section~\ref{section:the-dataset}, for the Small CNN Zoo dataset there is no relevant difference between the two and we do not really know which of them the GBM model predicts.
As a reference we also report the $\tau$ coefficients when using the train accuracy as a proxy for the test one (or vice versa).

All the numbers are significantly larger than zero, indicating that the predictor's ranking is far from being random.
The predictions seem to correlate slightly better with train accuracy than with test.
This hints that the predictors trained on the Small CNN Zoo may be using the train accuracy as a shortcut while predicting the test one.
The ranking coefficient between the train and test accuracies decreases with network size, which points to increasing overfitting.

To further verify that our findings hold up with other architectures, we show in Supplement~\ref{appendix:mlp-results} that our findings also hold for Multi-Layer Perceptrons.

\begin{table}[tb]
\vskip -0.1in
\caption{Kendall's $\tau$ coefficients between predictions and training/test accuracies of the ResNet32 models from the \mbox{DEMOGEN} dataset across different widths (columns).
Predictions are made with the GBM model trained on the CIFAR10-GS collection from the Small CNN Zoo using $\widetilde{{W}}_{\!L}^{4}$.
The GBM model is able to rank large ResNet models according to their accuracy, despite having been trained on small CNN architectures on different datasets.
}
\label{table:demogen}
\begin{center}
\begin{small}
\begin{sc}
\def\arraystretch{1.3}
\begin{tabular}{cccc}
\hline
Resnet-Width: & $\times 1$ & $\times 2$ & $\times 4$\\
\hline
Predictions vs Train & 0.30 & 0.62 & 0.50\\
Predictions vs Test & 0.28 & 0.59 & 0.32\\ \hline
Baseline: Train vs Test& 0.83 & 0.77 & 0.64\\
\hline
\end{tabular}
\end{sc}
\end{small}
\end{center}
\vskip -0.2in
\end{table}

\section{Conclusions and future directions}
We demonstrated that it is possible to predict the performance of a DNN using only its weights (or simple statistics thereof) as inputs. 
Surprisingly, these predictions are able to rank networks trained on unobserved natural image datasets/with different large architectures.
Whether these predictions can be reduced to simple human-interpretable rules
and whether they can be helpful to improve DNN training remains an important open question.
It also remains to be explored whether our findings transfer to domains outside of CNNs, e.g.\:to architectures commonly used in natural language understanding, reinforcement learning, or unsupervised applications.

Our work only used off-the-shelf regression algorithms (GBM and fully-connected DNNs) to predict the network accuracy using its weights.
In future it seems natural to try methods with stronger inductive biases.
For instance, using Deep Sets approach \cite{Zaheer2017} to account for the invariance of CNNs w.r.t.\:the order of the filters and channels could allow us to get even better performance in practical applications, or yield better insights.

We believe our findings open the door to a number of interesting further questions.
The idea that most neural network contain a highly efficient sub-network, the ``lottery ticket hypothesis'' \cite{Frankle2019}, recently gained a lot of attention. 
\citet{Morcos2019} show that these sub-networks transfer across tasks and datasets. 
An interesting avenue for future research would be to see if a trained classifier is able to identify these sub-networks (or other related properties) from the weights used to initialize a network. 

Finally, we share a large dataset of trained CNNs in hope that this will enable the community to further explore this interesting direction of research.

\section*{Acknowledgements}
We are thankful to 
Ibrahim Alabdulmohsin,
Iuliya Beloshapka,
Samy Bengio,
Lucas Beyer,
Alexey Dosovitskiy,
Pierre Foret,
Yiding Jiang,
Alexander Kolesnikov,
Dilip Krishnan,
Hossein Mobahi,
Shay Moran,
Behnam Neyshabur,
Sebastian Nowozin,
Paul Rubenstein,
Hanie Sedghi,
Jakob Uszkoreit,
and Scott Yak
for valuable discussions.

\bibliography{main}
\bibliographystyle{icml2020}

\newpage
\onecolumn
\appendix 

\section{Further details on the Small CNN Zoo dataset and experiments}
\label{appendix:small-cnn-zoo-specs}
This section contains details on the way the \textit{Small CNN Zoo} was generated and on the results of training the accuracy predictors reported in Tables 1 and 2 of the main text. 

\subsection{Base CNNs: architecture} 
All CNN models share the same architecture: 3 hidden convolutional layers with 16 filters each, followed by the global average pooling and the final dense layer. 
Dropout is applied to every convolutional layer. 
$\ell_2$-regularization is applied to all layers.
For exact details refer to the code at \url{https://github.com/google-research/google-research/tree/master/dnn_predict_accuracy}.

\subsection{Base CNNs: hyperparameters for training} 
\label{appendix:base-cnn-sweep}
For each dataset, we sample 30k different hyperparameter configurations of the CNN training:
\begin{itemize}
    \item
    \texttt{Optimizer} is chosen uniformly from one of the following: 
    vanilla SGD optimizer,
    Adam optimizer \cite{KL14}, and
    RMSProp optimizer;
    \item 
    \texttt{Learning rate} is sampled log-uniformly from
    $[5\times 10^{-4}, 5\times 10^{-2}]$;
    \item
    \texttt{$\ell_2$ regularization coefficient} is sampled log-uniformly from 
    $[10^{-8}, 10^{-2}]$;
    \item
    \texttt{Dropout rate} is sampled uniformly from $[0, 0.7]$;
    \item 
    \texttt{Variance of weight initializer} is sampled log-uniformly from $[10^{-3}, 0.5]$;
    \item
    \texttt{Type of weight initializer} is chosen uniformly from one of the following: 
    Xavier normal~\cite{glorot2010},
    He normal~\cite{he2015},
    orthogonal~\cite{saxe2014},
    normal, and truncated normal;
    \item
    Biases are initialized with zeros;
    \item
    \texttt{Activation function} is chosen uniformly from ReLu and hyperbolic tangent;
    \item
    \texttt{Fraction of training examples} to use is sampled uniformly from $\{0.1, 0.25, 0.5, 1.0\}$;
    \item
    We \emph{never} used same hyperparameter configuration with several different random seeds.
\end{itemize}

\subsection{Accuracy predictors: types of the models}
We use three types of predictors: logit-linear models, 
gradient boosted machine using desicion trees (GBM),
and fully-connected ReLu networks (DNN).

The {\bf logit-linear model} $W\mapsto \sigma(\langle W, \theta\rangle + b)$ takes the output of the linear model and transforms it with the sigmoid function $\sigma(z):=(1 + e^{-z})^{-1}$. 
Here $\langle x, y\rangle$ denotes the inner product.
We use a logit-linear (instead of plain linear) model because the targets (test accuracies) are in $[0,1]$ and in preliminary experiments we did not observe a linear model that achieved a better performance. 
We train the parameters $\theta$ and $b$ with 
mini-batch SGD/Adam varying the learning rate, batch size, initialization, and $\ell_2$-regularization.

We use LightGBM \cite{Ke2017} to train the {\bf GBM model} and vary the number of leaves and maximum depth of the trees, the learning rate, $\ell_1$ and $\ell_2$ regularization, and parameters for the features/examples subsampling.

We use a feed-forward fully-connected architecture for the {\bf DNN model} with ReLU activations and sigmoid transform.
We train with mini-batch SGD/Adam varying the learning rate, number of layers and their width, $\ell_2$-regularization, initialization type and variance, and batch size.

\subsection{Accuracy predictors: hyperparameters for training}
\label{subsection:predictor-hyper}
For each of the 3 types of predictors and each of the 4 CNN collections we perform hyperparameter selection by evaluating 1k
unique configurations:
\begin{itemize}
    \item For the GBM accuracy predictor we use the following protocol. Refer to the Light-GBM documentation for the exact meaning of the parameters:
    \begin{itemize}
        \item \texttt{num\_leaves} is sampled uniformly from $[20, 10^4]$;
        \item \texttt{max\_depth} is sampled uniformly from $[5, 15]$;
        \item \texttt{learning\_rate} is sampled log-uniformly from $[10^{-2}, 10^{-1}]$;
        \item \texttt{max\_bin} is sampled uniformly from $\{2^6 - 1, 2^7 - 1, 2^8 - 1,\}$;
        \item \texttt{min\_child\_weight} is sampled uniformly from $\{1, 2, 3, 4, 5\}$;
        \item \texttt{reg\_lambda} is sampled uniformly from $[10^{-3}, 100]$;
        \item \texttt{reg\_alpha} is sampled uniformly from $[10^{-6}, 5]$;
        \item \texttt{subsample} is sampled uniformly from $\{0.1, 0.2, \dots, 0.9, 1\}$;
        \item \texttt{subsample\_freq} is set to 1;
        \item \texttt{colsample\_bytree}
        for the high dimensional inputs 
        (all weights $W$, weights of the second and third convolutional layers $W^2$ and $W^3$, and concatenation of all weights with the hyperparameters $(\lambda, W)$) is sampled log-uniformly from $[10^{-2}, 10^{-1}]$, for lower dimensional inputs is sampled uniformly from $[0.7, 1]$;
    \end{itemize}
    \item For the DNN accuracy predictor we use the following protocol:
    \begin{itemize}
        \item \texttt{Number of layers} is sampled uniformly from $\{3,4,\dots,9\}$;
        \item \texttt{Number of units} is sampled uniformly from $\{256, 257,\dots, 511\}$;
        \item ReLu activation is used for all models;
        \item \texttt{Dropout rate} is sampled uniformly from $[0, 0.2]$;
        \item \texttt{$\ell_2$-regularization coefficient} is sampled log-uniformly from $[10^{-8}, 10^{-3}]$;
        \item \texttt{Learning rate} is sampled log-uniformly from $[10^{-3}, 0.5]$;
        \item \texttt{Variance of weight initializer} is sampled log-uniformly from $[10^{-3}, 0.1]$;
        \item Optimizer is chosen randomly from Adam and SGD;
        \item 
        \texttt{Batch size} is sampled uniformly from $\{64, 128, 256, 512\}$;
        \item
        Biases are initialized with zeros;
        \item
        \texttt{Type of weight initializer} is chosen uniformly from one of the following: 
        Xavier normal \cite{glorot2010},
        He normal \cite{he2015},
        orthogonal \cite{saxe2014},
        normal, and truncated normal;
        \item Sigmoid transform is applied to the final layer output.
    \end{itemize}
    \item For the logit-linear predictor we use the same protocol as for DNN predictor, while setting \texttt{Number of layers} to zero and applying $\ell_2$ regularization to the final dense layer.
\end{itemize}

\subsection{Accuracy predictors: detailed empirical results}
\label{appendix:predictors-detailed}

Tables \ref{table:bigtable-r2} and \ref{table:bigtable-mse} contain both $R^2$ scores and MSE values for all three types of predictors trained on all four CNN collections.
Standard deviations capture the variability when training the predictors on three folds of the cross-validation.
Every entry in the Tables \ref{table:bigtable-r2} and \ref{table:bigtable-mse} is obtained by evaluating 1k hyperparameter configurations of the accuracy predictor (as described in Section \ref{subsection:predictor-hyper}) and picking the best one using 3-fold cross validation. 
Then the best configuration is evaluated on the holdout test split of the CNN collection.
The resulting numbers are reported in the tables.

Tables \ref{table:bigtable-r2} and \ref{table:bigtable-mse} do not contain results for several input types, including the ones based on norms $W^{\ell_1}_L$ and $W^{\ell_2}_L$, 
on weights statistics computed for subsets of layers $\widetilde{W}_L^4$ and $\widetilde{W}_L^{1, 4}$,
and on learning rate $\lambda_{\mathrm{LR}}$.
Results for these input types are reported in Tables \ref{table:remaining-r2} and \ref{table:remaining-mse}.

\begin{table}
\caption{$R^2$ (together with standard deviations) for predicting test accuracies of CNNs trained on various datasets (blocks) with various models (rows) using different input features ({columns}).
S.t.d.\:capture the variability when training the models on 3 different folds of the cross-validation.
``Lin'' refers to the logit-linear model.
See main text for the descriptions of input features.}
\label{table:bigtable-r2}
\begin{center}
\resizebox{\textwidth}{!}{%
\begin{tiny}%
\def\arraystretch{1.5}%
\begin{tabular}{cccccccccc}
\hline
& $W^1$ & ${W^2}$ & ${W^3}$ & ${W^4}$ & ${W}$ & $\lambda$ & $\lambda, W$ & $\widetilde{{W}}$ & $\widetilde{{W}}_{L}$ \\ \hline
\multicolumn{10}{c}{MNIST}\\ \hline
Lin & .716$\pm$.001 & .634$\pm$.002 & .659$\pm$.003 & .808$\pm$.001 & .847$\pm$.003 & .692$\pm$.001 & .874$\pm$.002 & .780$\pm$.001 & .920$\pm$.000 \\
GBM & .977$\pm$.001 & .966$\pm$.001 & .969$\pm$.001 & .987$\pm$.000 & .988$\pm$.000 & .918$\pm$.002 & .990$\pm$.000 & .953$\pm$.001 & .993$\pm$.000 \\
DNN & .978$\pm$.000 & .969$\pm$.000 & .975$\pm$.001 & .989$\pm$.001 & .980$\pm$.001 & .898$\pm$.003 & .979$\pm$.001 & .948$\pm$.001 & .993$\pm$.000 \\ \hline
\multicolumn{10}{c}{Fashion MNIST}\\ \hline
Lin & .500$\pm$.002 & .423$\pm$.004 & .450$\pm$.006 & .699$\pm$.000 & .733$\pm$.003 & .614$\pm$.001 & .793$\pm$.003 & .516$\pm$.002 & .804$\pm$.001 \\
GBM & .982$\pm$.000 & .975$\pm$.001 & .973$\pm$.000 & .989$\pm$.000 & .989$\pm$.000 & .924$\pm$.000 & .991$\pm$.000 & .955$\pm$.001 & .993$\pm$.000 \\
DNN & .982$\pm$.001 & .973$\pm$.000 & .976$\pm$.000 & .989$\pm$.001 & .980$\pm$.001 & .888$\pm$.003 & .983$\pm$.000 & .938$\pm$.009 & .992$\pm$.000 \\ \hline
\multicolumn{10}{c}{CIFAR10-GS}\\ \hline
Lin & .361$\pm$.004 & .404$\pm$.006 & .428$\pm$.004 & .707$\pm$.000 & .662$\pm$.002 & .685$\pm$.001 & .735$\pm$.004 & .186$\pm$.001 & .727$\pm$.002 \\
GBM & .959$\pm$.000 & .926$\pm$.001 & .928$\pm$.000 & .969$\pm$.000 & .970$\pm$.000 & .934$\pm$.001 & .979$\pm$.000 & .914$\pm$.000 & .984$\pm$.000 \\
DNN & .959$\pm$.000 & .929$\pm$.001 & .930$\pm$.002 & .968$\pm$.001 & .954$\pm$.001 & .903$\pm$.005 & .955$\pm$.001 & .897$\pm$.005 & .980$\pm$.001 \\ \hline
\multicolumn{10}{c}{SVHN-GS}\\ \hline
Lin & .451$\pm$.003 & .545$\pm$.000 & .566$\pm$.003 & .802$\pm$.002 & .786$\pm$.003 & .636$\pm$.001 & .814$\pm$.002 & .384$\pm$.001 & .852$\pm$.001 \\
GBM & .936$\pm$.000 & .895$\pm$.001 & .900$\pm$.000 & .967$\pm$.000 & .971$\pm$.000 & .935$\pm$.001 & .978$\pm$.000 & .908$\pm$.000 & .986$\pm$.000 \\
DNN & .935$\pm$.001 & .876$\pm$.001 & .890$\pm$.002 & .973$\pm$.000 & .931$\pm$.003 & .879$\pm$.022 & .934$\pm$.001 & .904$\pm$.002 & .985$\pm$.001 \\ \hline
\end{tabular}
\end{tiny}%
}
\end{center}
\end{table}

\begin{table}
\caption{MSE (together with standard deviations) for predicting test accuracies of CNNs trained on various datasets (blocks) with various models (rows) using different input features ({columns}).
S.t.d.\:capture the variability when training the models on 3 different folds of the cross-validation.
``Lin'' refers to the logit-linear model.
See main text for the descriptions of input features.}
\label{table:bigtable-mse}
\begin{center}
\resizebox{\textwidth}{!}{%
\begin{tiny}
\def\arraystretch{1.5}
\begin{tabular}{cccccccccc}
\hline
& $W^1$ & ${W^2}$ & ${W^3}$ & ${W^4}$ & ${W}$ & $\lambda$ & $\lambda, W$ & $\widetilde{{W}}$ & $\widetilde{{W}}_{L}$ \\ \hline
\multicolumn{10}{c}{MNIST}\\ \hline
Lin & .025$\pm$.000 & .033$\pm$.000 & .030$\pm$.000 & .017$\pm$.000 & .014$\pm$.000 & .028$\pm$.000 & .011$\pm$.000 & .020$\pm$.000 & .007$\pm$.000 \\
GBM & .002$\pm$.000 & .003$\pm$.000 & .003$\pm$.000 & .001$\pm$.000 & .001$\pm$.000 & .007$\pm$.000 & .001$\pm$.000 & .004$\pm$.000 & .001$\pm$.000 \\
DNN & .002$\pm$.000 & .003$\pm$.000 & .002$\pm$.000 & .001$\pm$.000 & .002$\pm$.000 & .009$\pm$.000 & .002$\pm$.000 & .005$\pm$.000 & .001$\pm$.000 \\ \hline
\multicolumn{10}{c}{Fashion MNIST}\\ \hline
Lin & .026$\pm$.000 & .030$\pm$.000 & .029$\pm$.000 & .016$\pm$.000 & .014$\pm$.000 & .020$\pm$.000 & .011$\pm$.000 & .025$\pm$.000 & .010$\pm$.000 \\
GBM & .001$\pm$.000 & .001$\pm$.000 & .001$\pm$.000 & .001$\pm$.000 & .001$\pm$.000 & .004$\pm$.000 & .000$\pm$.000 & .002$\pm$.000 & .000$\pm$.000 \\
DNN & .001$\pm$.000 & .001$\pm$.000 & .001$\pm$.000 & .001$\pm$.000 & .001$\pm$.000 & .006$\pm$.001 & .001$\pm$.000 & .003$\pm$.000 & .000$\pm$.000 \\ \hline
\multicolumn{10}{c}{CIFAR10-GS}\\ \hline
Lin & .009$\pm$.000 & .008$\pm$.000 & .008$\pm$.000 & .004$\pm$.000 & .005$\pm$.000 & .004$\pm$.000 & .004$\pm$.000 & .011$\pm$.000 & .004$\pm$.000 \\
GBM & .001$\pm$.000 & .001$\pm$.000 & .001$\pm$.000 & .000$\pm$.000 & .000$\pm$.000 & .001$\pm$.000 & .000$\pm$.000 & .001$\pm$.000 & .000$\pm$.000 \\
DNN & .001$\pm$.000 & .001$\pm$.000 & .001$\pm$.000 & .000$\pm$.000 & .001$\pm$.000 & .001$\pm$.000 & .001$\pm$.000 & .001$\pm$.000 & .000$\pm$.000 \\ \hline
\multicolumn{10}{c}{SVHN-GS}\\ \hline
Lin & .012$\pm$.000 & .010$\pm$.000 & .009$\pm$.000 & .004$\pm$.000 & .005$\pm$.000 & .008$\pm$.000 & .004$\pm$.000 & .013$\pm$.000 & .003$\pm$.000 \\
GBM & .001$\pm$.000 & .002$\pm$.000 & .002$\pm$.000 & .001$\pm$.000 & .001$\pm$.000 & .001$\pm$.000 & .000$\pm$.000 & .002$\pm$.000 & .000$\pm$.000 \\
DNN & .001$\pm$.000 & .003$\pm$.000 & .002$\pm$.000 & .001$\pm$.000 & .001$\pm$.000 & .003$\pm$.000 & .001$\pm$.000 & .002$\pm$.000 & .000$\pm$.000 \\ \hline
\end{tabular}
\end{tiny}%
}
\end{center}
\end{table}

\begin{table}
\caption{$R^2$ (together with standard deviations) for predicting test accuracies of CNNs trained on various datasets (blocks) with various models (rows) using different input features ({columns}).
S.t.d.\:capture the variability when training the models on 3 different folds of the cross-validation.}
\label{table:remaining-r2}
\begin{center}
\begin{small}
\def\arraystretch{1.5}
\begin{tabular}{cccccc}
\hline
& $W^{\ell_1}_L$ & $W^{\ell_2}_L$ & $\widetilde{W}_L^{4}$ & $\widetilde{W}_L^{1, 4}$ & $\lambda_{\mathrm{LR}}$ \\ \hline
\multicolumn{6}{c}{MNIST}\\ \hline
Lin & 0.835 $\pm$ 0.002 & 0.848 $\pm$ 0.000 & 0.842 $\pm$ 0.000 & 0.902 $\pm$ 0.000 & 0.025 $\pm$ 0.000 \\
GBM & 0.983 $\pm$ 0.000 & 0.981 $\pm$ 0.000 & 0.973 $\pm$ 0.000 & 0.989 $\pm$ 0.000 & 0.024 $\pm$ 0.000 \\
DNN & 0.959 $\pm$ 0.004 & 0.980 $\pm$ 0.002 & 0.973 $\pm$ 0.000 & 0.988 $\pm$ 0.001 & 0.026 $\pm$ 0.000 \\ \hline
\multicolumn{6}{c}{Fashion MNIST}\\ \hline
Lin & 0.716 $\pm$ 0.013 & 0.730 $\pm$ 0.001 & 0.713 $\pm$ 0.000 & 0.775 $\pm$ 0.001 & 0.033 $\pm$ 0.000 \\
GBM & 0.982 $\pm$ 0.000 & 0.983 $\pm$ 0.000 & 0.976 $\pm$ 0.000 & 0.989 $\pm$ 0.000 & 0.035 $\pm$ 0.000 \\
DNN & 0.952 $\pm$ 0.006 & 0.981 $\pm$ 0.001 & 0.975 $\pm$ 0.001 & 0.988 $\pm$ 0.001 & 0.036 $\pm$ 0.000 \\ \hline
\multicolumn{6}{c}{CIFAR10-GS}\\ \hline
Lin & 0.479 $\pm$ 0.036 & 0.542 $\pm$ 0.001 & 0.605 $\pm$ 0.001 & 0.633 $\pm$ 0.003 & 0.003 $\pm$ 0.001 \\
GBM & 0.960 $\pm$ 0.000 & 0.960 $\pm$ 0.001 & 0.941 $\pm$ 0.001 & 0.971 $\pm$ 0.000 & 0.015 $\pm$ 0.000 \\
DNN & 0.927 $\pm$ 0.006 & 0.950 $\pm$ 0.001 & 0.933 $\pm$ 0.002 & 0.968 $\pm$ 0.001 & 0.014 $\pm$ 0.003 \\ \hline
\multicolumn{6}{c}{SVHN-GS}\\ \hline
Lin & 0.629 $\pm$ 0.005 & 0.700 $\pm$ 0.000 & 0.755 $\pm$ 0.001 & 0.779 $\pm$ 0.001 & 0.000 $\pm$ 0.000 \\
GBM & 0.967 $\pm$ 0.000 & 0.971 $\pm$ 0.000 & 0.946 $\pm$ 0.001 & 0.971 $\pm$ 0.000 & 0.034 $\pm$ 0.001 \\
DNN & 0.945 $\pm$ 0.001 & 0.971 $\pm$ 0.001 & 0.946 $\pm$ 0.001 & 0.971 $\pm$ 0.000 & 0.034 $\pm$ 0.000 \\ \hline
\end{tabular}
\end{small}
\end{center}
\end{table}

\begin{table}
\caption{MSE (together with standard deviations) for predicting test accuracies of CNNs trained on various datasets (blocks) with various models (rows) using different input features ({columns}).
S.t.d.\:capture the variability when training the models on 3 different folds of the cross-validation.}
\label{table:remaining-mse}
\begin{center}
\begin{small}
\def\arraystretch{1.5}
\begin{tabular}{cccccc}
\hline
& $W^{\ell_1}_L$ & $W^{\ell_2}_L$ & $\widetilde{W}_L^{4}$ & $\widetilde{W}_L^{1, 4}$ & $\lambda_{\mathrm{LR}}$ \\ \hline
\multicolumn{6}{c}{MNIST}\\ \hline
Lin & 0.015 $\pm$ 0.000 & 0.014 $\pm$ 0.000 & 0.014 $\pm$ 0.000 & 0.009 $\pm$ 0.000 & 0.087 $\pm$ 0.000 \\
GBM & 0.002 $\pm$ 0.000 & 0.002 $\pm$ 0.000 & 0.002 $\pm$ 0.000 & 0.001 $\pm$ 0.000 & 0.087 $\pm$ 0.000 \\
DNN & 0.004 $\pm$ 0.000 & 0.002 $\pm$ 0.000 & 0.002 $\pm$ 0.000 & 0.001 $\pm$ 0.000 & 0.087 $\pm$ 0.000 \\ \hline
\multicolumn{6}{c}{Fashion MNIST}\\ \hline
Lin & 0.015 $\pm$ 0.001 & 0.014 $\pm$ 0.000 & 0.015 $\pm$ 0.000 & 0.012 $\pm$ 0.000 & 0.051 $\pm$ 0.000 \\
GBM & 0.001 $\pm$ 0.000 & 0.001 $\pm$ 0.000 & 0.001 $\pm$ 0.000 & 0.001 $\pm$ 0.000 & 0.051 $\pm$ 0.000 \\
DNN & 0.003 $\pm$ 0.000 & 0.001 $\pm$ 0.000 & 0.001 $\pm$ 0.000 & 0.001 $\pm$ 0.000 & 0.051 $\pm$ 0.000 \\ \hline
\multicolumn{6}{c}{CIFAR10-GS}\\ \hline
Lin & 0.007 $\pm$ 0.000 & 0.006 $\pm$ 0.000 & 0.005 $\pm$ 0.000 & 0.005 $\pm$ 0.000 & 0.014 $\pm$ 0.000 \\
GBM & 0.001 $\pm$ 0.000 & 0.001 $\pm$ 0.000 & 0.001 $\pm$ 0.000 & 0.000 $\pm$ 0.000 & 0.013 $\pm$ 0.000 \\
DNN & 0.001 $\pm$ 0.000 & 0.001 $\pm$ 0.000 & 0.001 $\pm$ 0.000 & 0.000 $\pm$ 0.000 & 0.014 $\pm$ 0.000 \\ \hline
\multicolumn{6}{c}{SVHN-GS}\\ \hline
Lin & 0.008 $\pm$ 0.000 & 0.006 $\pm$ 0.000 & 0.005 $\pm$ 0.000 & 0.005 $\pm$ 0.000 & 0.021 $\pm$ 0.000 \\
GBM & 0.001 $\pm$ 0.000 & 0.001 $\pm$ 0.000 & 0.001 $\pm$ 0.000 & 0.001 $\pm$ 0.000 & 0.021 $\pm$ 0.000 \\
DNN & 0.001 $\pm$ 0.000 & 0.001 $\pm$ 0.000 & 0.001 $\pm$ 0.000 & 0.001 $\pm$ 0.000 & 0.021 $\pm$ 0.000 \\ \hline
\end{tabular}
\end{small}
\end{center}
\end{table}

\section{GBM importance plots}
\label{appendix:gbm-importances}
\begin{figure}[tbp]
\begin{center}
\caption{Light-GBM feature importance values based on number of times the feature appeared in the trees. 
    Four plots correspond to GBM predictors  trained on four CNN collections using entire weight vectors $W$ as inputs.
    ``L'' in feature names refer to the layer, ``W'' to the (filter) weights,
    ``B'' to the biases.
    For example, ``L4-B7'' is the 7th bias parameter of the final dense layer and ``L1-W123'' is the 123rd filter weight parameter of the first convolutional layer.}
    \centerline{\includegraphics[width=\linewidth]{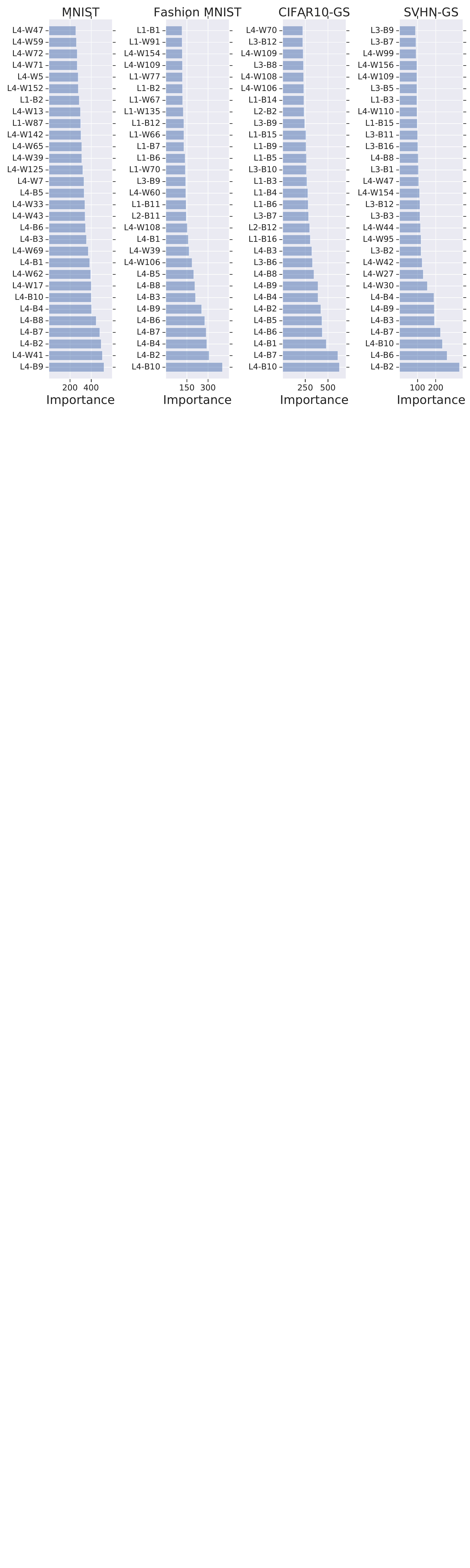}}
    \label{fig:importances}
\end{center}
\vskip -.4in
\end{figure}

Figure \ref{fig:importances} presents importance values for various entries of the weight vector $W$ when training the GBM accuracy predictor.
Importance values reported in the figure are based on the number of times a single feature (a particular entry of the vector $W$ in our case) was chosen in the nodes of the trees. 
Higher numbers correspond to more important (\emph{more frequently used}) features.
We see that all four models make extensive use of parameters of the final dense layer.
Among those, biases seem to be slightly more important than weights.

\section{Permutation and scale invariance}
\label{appendix:invariance-study}
This section contains the results of a study on how accuracy estimator’s predictions change as we modify its inputs.
For a given accuracy predictor $\hat{F}$ trained using weight vectors $W$ as inputs we test several ways of modifying its inputs $W\mapsto \varphi(W)$:
\begin{enumerate}
\item
Globally permuting the elements of $W$;
\item
Permuting the order of parameters within each layer of $W$;
\item
Permuting the order of parameters within all three convolutional layers of $W$;
\item
Permuting the order of parameters in the final dense layer;
\item
Multiplying all elements of $W$ by a constant $c > 0$.
\end{enumerate}
For every type of permutation we try two options: (a) permuting biases and weights jointly, allowing them to mix and (b) permuting biases and weights separately, without mixing them.

We test these modifications with the GBM predictor $\hat{F}$ trained using weight vectors $W$ as inputs on the CIFAR10-GS CNN collection, which has the $R^2$ score of 0.97.
We use uniformly sampled random permutations and scale factors $c\in\{10^{-3}, 10^{-1}, 2, 10, 100\}$.
For every type of modification $\varphi$ we take the absolute differences $|\hat{F}\bigl(\varphi(W)\bigr) - \hat{F}(W)|$ between the predictions on the modified and original CNNs respectively.
Then we average them across 1000 CNNs $W$ from the test split of the CIFAR10-GS CNN collection.
Results are reported in Table \ref{table:SCNNZ-invariances}.

\begin{table}[htb]
\caption{Sensitivity of the GBM predictor trained using weights $W$ on the CIFAR10-GS collection w.r.t.\,various modifications of its inputs. 
}
\label{table:SCNNZ-invariances}
\vskip 0.2in
\begin{center}
\begin{small}
\begin{sc}
\def\arraystretch{1.5}
\begin{tabular}{lc}
\hline
Modification $\varphi$ & MAD\\
\hline
Scale, $c=10^{-3}$ & 0.1353\\
Scale, $c=10^{-1}$ & 0.1144\\
Global permutation & 0.1100\\
Permuting within all layers (mixing b.\:and w.) & 0.0893\\
Permuting within final layer (mixing b.\:and w.) & 0.0680\\
Permuting within all layers (not mixing b.\:and w.) & 0.0671\\
Permuting within final layer (not mixing b.\:and w.) & 0.0585\\
Scale, $c=100$ & 0.0470\\
Scale, $c=10$ & 0.0465\\
Scale, $c=2$ & 0.0401\\
Permuting within first 3 layers (mixing b.\:and w.) & 0.0311\\
Permuting within first 3 layers (not mixing b.\:and w.) & 0.0133\\
\hline
\end{tabular}
\end{sc}
\end{small}
\end{center}
\vskip -0.2in
\end{table}

\section{Detailed results on the transfer experiments}
\label{appendix:transfer-plots}
\begin{figure}[tbp]
\begin{center}
    \centerline{\includegraphics[width=\linewidth]{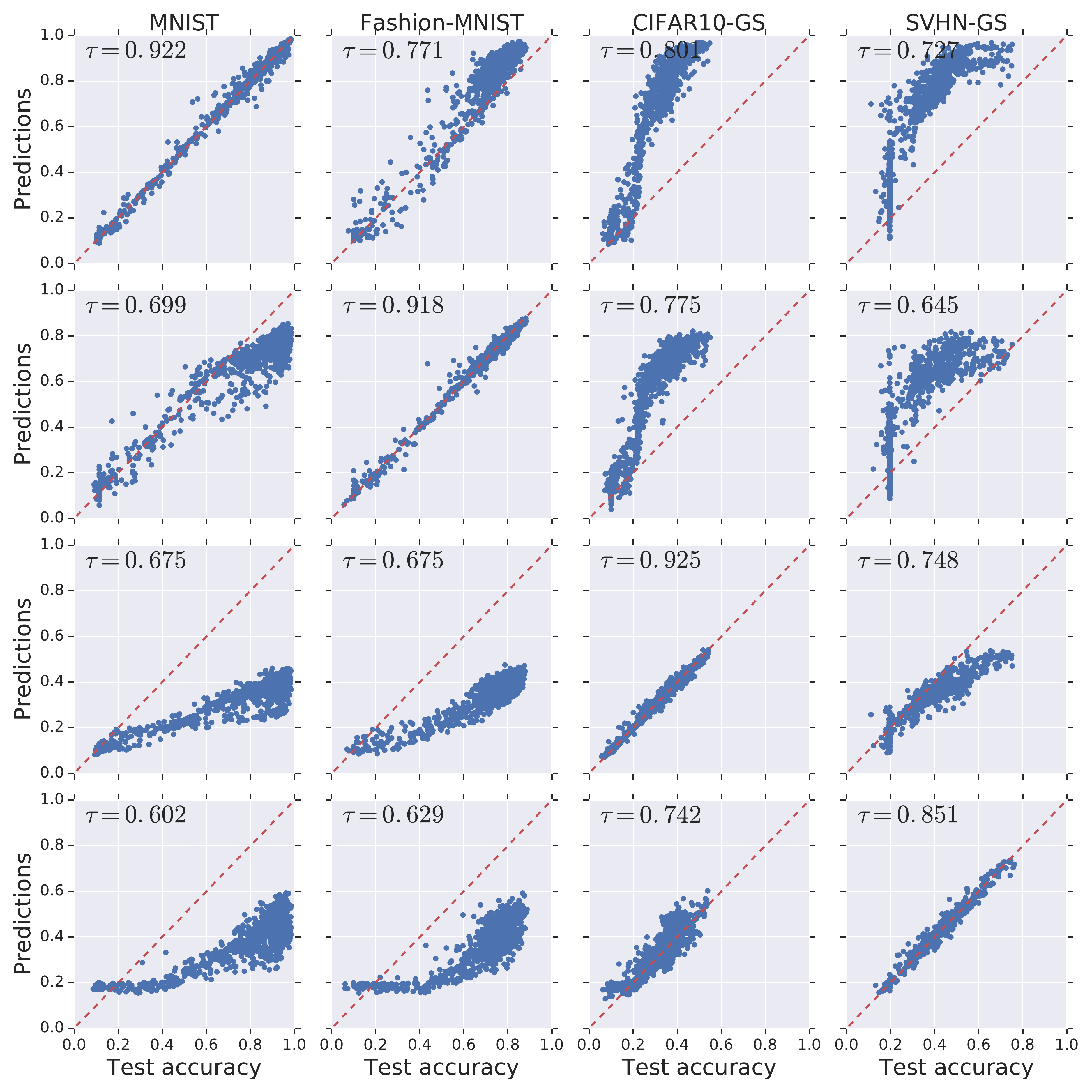}}
    \caption{Distribution of true/predicted test accuracies for networks from different CNN collections (columns) together with Kendall's $\tau$ coefficients. 
    Predictions were made with the GBM models trained on different CNN collections (rows, same order) using $\wstats$.}
    \label{fig:transfer-datasets}
\end{center}
\end{figure}

Figure \ref{fig:transfer-datasets} contains the results for all possible transfer experiments performed on Small CNN Zoo as described in Section~\ref{section:smallcnnzoo-transfer}.
Diagonal plots correspond to the holdout test evaluation of four GBM models.

\clearpage 
\section{Results on Multi-Layer Perceptrons}
\label{appendix:mlp-results}
To verify that our results do not only apply to CNNs, we performed experiments on fully connected Multi-Layer Perceptrons (MLPs). We trained 10k MLPs each on CIFAR10 and SVHN, using the same hyperparameters as in the Small CNN Zoo (see Section~\ref{appendix:base-cnn-sweep}), except that we also sampled the number of hidden units in each layer to be either 8, 16, 32 or 64. This gave us four different neural network sizes for each dataset. To save computation time and verify that our observations also hold with different estimators, we used a Random Forest estimator with 32 trees in all of the experiments. The estimator was trained on networks from one specific dataset and hidden-unit size, and evaluated on all other settings. We used the same weight statistics $\wstats$ as in the main text as input features. Table~\ref{table:mlp-results} shows the results of this experiment. We then verified that these results transfer across network architectures and datasets: The resulting Kendall's $\tau$ coefficients are listed in Table~\ref{table:mlp-transfer-results}. Together, these results confirm our CNN findings, namely that it is possible to predict the performance of an MLP based on its weights, and that this prediction transfers across models of different sizes as well as across datasets.

\begin{table}
\caption{Kendall's $\tau$ coefficients between predictions and test-accuracies on networks of different sizes and/or datasets. 80\,\% of the data was used for training, and 20\,\% for evaluation. Subscript $u$ indicates that the estimator was trained on networks with hidden layers of size $u$.}
\label{table:mlp-results}
\begin{center}
\begin{small}
\def\arraystretch{1.5}
\begin{tabular}{rrrrrrrr}
CIFAR10$_{8}$ & CIFAR10$_{16}$ & CIFAR10$_{32}$ & CIFAR10$_{64}$ & SVHN$_{8}$ & SVHN$_{16}$ & SVHN$_{32}$ & SVHN$_{64}$\\
\hline
0.924 & 0.939 & 0.925 & 0.942 & 0.905 & 0.917 & 0.925 & 0.924
\end{tabular}
\end{small}
\end{center}
\end{table}

\begin{table}
\caption{Kendall's $\tau$ coefficients between predictions and test-accuracies on MLPs of different sizes and/or datasets. For each row, we show the results of training on a given dataset/MLP-size and evaluating on all other sizes. Subscript $u$ indicates that the estimator was trained on networks with hidden layers of size $u$.}
\label{table:mlp-transfer-results}
\begin{center}
\begin{small}
\def\arraystretch{1.5}
\begin{tabular}{l|rrrrrrrr}
\hline
 & CIFAR10$_{8}$ & CIFAR10$_{16}$ & CIFAR10$_{32}$ & CIFAR10$_{64}$ & SVHN$_{8}$ & SVHN$_{16}$ & SVHN$_{32}$ & SVHN$_{64}$\\
\hline
CIFAR10$_{8}$	&  --  & 0.757 & 0.651 & 0.555 & 0.582 & 0.614 & 0.607 & 0.574\\
CIFAR10$_{16}$	& 0.759 &  --  & 0.758 & 0.653 & 0.504 & 0.608 & 0.613 & 0.599\\
CIFAR10$_{32}$	& 0.624 & 0.742 &  --  & 0.787 & 0.496 & 0.570 & 0.633 & 0.652\\
CIFAR10$_{64}$	& 0.559 & 0.655 & 0.771 &  --  & 0.461 & 0.554 & 0.599 & 0.633\\
SVHN$_{8}$	& 0.489 & 0.513 & 0.500 & 0.482 &  --  & 0.805 & 0.715 & 0.627\\
SVHN$_{16}$	& 0.532 & 0.558 & 0.568 & 0.534 & 0.772 &  --  & 0.798 & 0.719\\
SVHN$_{32}$	& 0.548 & 0.602 & 0.630 & 0.599 & 0.720 & 0.809 &  --  & 0.801\\
SVHN$_{64}$	& 0.559 & 0.655 & 0.689 & 0.654 & 0.702 & 0.752 & 0.801 &  -- \\
\end{tabular}
\end{small}
\end{center}
\end{table}

\end{document}